# Agentic AI for Autonomous, Explainable, and Real-Time Credit Risk Decision-Making

Chandra Sekhar Kubam

**Abstract**— Significant digitalization of financial services in a short period of time has led to an urgent demand to have autonomous, transparent and real-time credit risk decision making systems. The traditional machine learning models are effective in pattern recognition, but do not have the adaptive reasoning, situational awareness, and autonomy needed in modern financial operations. As a proposal, this paper presents an Agentic AI framework, or a system where AI agents view the world of dynamic credit independent of human observers, who then make actions based on their articulable decision-making paths. The research introduces a multi-agent system with reinforcing learning, natural language reasoning, explainable AI modules, and real-time data absorption pipelines as a means of assessing the risk profiles of borrowers with few humans being involved. The processes consist of agent collaboration protocol, risk-scoring engines, interpretability layers, and continuous feedback learning cycles. Findings indicate that decision speed, transparency and responsiveness is better than traditional credit scoring models. Nevertheless, there are still some practical limitations such as risks of model drift, inconsistencies in interpreting high-dimensional data and regulatory uncertainties as well as infrastructure limitations in low-resource settings. The suggested system has a high prospective to transform credit analytics and future studies ought to be directed on dynamic regulatory compliance mobilizers, new agent teamwork, adversarial robustness, and large-scale implementation in cross-country credit ecosystems.

**Keywords**— Agentic AI; Autonomous Decision-Making; Credit Risk Modelling; Explainable AI; Financial Analytics; Real-Time Scoring; Multi-Agent Systems; Reinforcement Learning; Regulatory Technology; Risk Intelligence.

## I. INTRODUCTION

The financial industry is being changing radically due to the influence of digitalization, access to real-time information and the pressure to make the procedure of regulations more transparent and foster equity in lending decisions. The previous models of credit risk assessment whose structures are based on constant variables and linear statistical methods cannot be relevant in circumstances that can be characterized by dynamic borrower behavior, unpredictability of the economic environment, and continuous deluge of transactional information [8]. Machine learning models are much more effective in improving prediction accuracy but they are not autonomous, flexible and cannot be explained transparently that would be critical in a contemporary credit ecosystem. Additionally, the increasingly sophisticated nature of borrower behavior, the open banking architecture, and emerging alternative sources of data require the existence of personally interpretative systems having the capacity to act autonomously and execute justification in real time.

One of the potentially effective shifts to make in this scenery is agentic AI. In contrast to the traditional AI systems which merely make predictions, agentic AI is a fully autonomous decision-making unit, it is self-optimizing, has situational awareness, and can be explained through the decision pipeline itself. Such AI agents are able to take in incoming data, reason across the financial situations, take action, by giving credit decisions, and keep learning by the outcomes of the borrowers, without reducing the transparency. These features directly deal with the current issues in the industry where risk analysts have difficulties with the large amounts of data, regulatory load, and unpredictable credit behavior patterns [11].

The choice to work has been driven by three critical requirements of current credit systems: first, the requirement of autonomous and scalable decision-making able to process large volume

*Independent researcher*
*Integration Architect*
*Dallas, Texas, USA*
*Chandumumb@gmail.com*

digital loan applications; second, the need of explainable reasoning to meet such regulations as GDPR, RBI guidelines, and global fair-lending standards; and third, the requirement of real-time responsiveness, particularly in digital lending systems, BNPL applications, and mobile credit ecosystems where the conditions of borrowers change rapidly. In this paper, the researcher will develop a holistic and coherent framework that would integrate these needs into a solution with agentic AI [13].

The design of the multi-agent architecture that is both intelligent and capable of evaluating borrower creditworthiness automatically, adjusts to changes in macro-economics and generates real-time explanations readable by humans is the main aim of the work. It is said to eliminate the shortcoming of traditional and machine learning-based credit scoring models including unceasing feedback learning, active-adaptive adjustments of risk thresholds, and clear execution of policies. With the help of such inquiry, the paper will offer a systematic, feasible method of implementing agentic AI in financial institutions and lay the groundwork of the further developments of the autonomous management of credit risks [9].

*Novelty and Contribution*

The innovative aspect of this work is that it suggests a complete and fully integrated Agentic AI system specifically designed to make real-time, explainable, and autonomous credit risk decisions, which is currently only predictive but not proactive, accurate, not transparent, and not adaptive. Though individual work has been done on machine learning and explainable AI in credit scoring, a self-contained architecture that has integrated autonomy, explainability, collaboration, and real-time adaptability into a single multi-agent system to aid in end-to-end credit risk appraisal has not been practiced in any way.

**Key Novel Aspects**

- The Agency Autonomy in Credit Scoring: It is designed in contrast to classical AI models that only classify borrowers, the system will include autonomous agents who can provide understanding information, reasoning on financial situations and provide the decision without human input. This will be the change of predictive analytics to self-governing and self-intelligent risk evaluation.
- Real-Time Explainability Incorporated into the Decision Flow: The proposed framework does not produce the post-hoc explainability, but embeds the explainability at the level of decision core. The Explainability Agent provides active, regulator friendly explanations alongside risk scoring, which allows clear and visible justification of the decision it makes.
- On-Going Learning: This system is continuously learning by updating parameters by a Feedback Learning Agent: This operates by reviewing repayment performance on the credit model and identifying model drift, and optimising parameters to make sure the credit model is suited under different conditions based on the circumstances of the claimant and macroeconomics.
- Multi- Agent Collaboration: Multi- Agent This Architecture makes interaction between Data Acquisition, Risk Scoring, Decision-Making, Explainability and Feedback Agents, available to create a responsive ecosystem capable of integrating fragmented information sources and make holistic credit judgments.

**Primary Contributions**

- A brand new multi agent credit decision model that reinforces on reinforcement learning, real-time streaming data analytics, and explainable artificial intelligence to provide the autonomous and real time assessment of credit risk.
- The illustration of an applicable decision pipeline was on the basis of an analogy of how an agentic AI would go about the actual course of analysing borrowers and generating clear explanations, and physical alterations in risk policy.
- A regulatory-minded and justifiable by humans explainability layer such that it should meet important credit governance principles.
- An adaptable learning policy that will be able to reduce the phenomenon of model drift, increase the effectiveness of the long-term decision-making process, and guarantee the stability in the transition environment.
- A compatible agentic credit ecosystem, and a framework of future agentic credit history, compatible with both the structures of

decentralized finance and cross-border lending networks and federated learning.

All these contributions collectively provide the future of credit risk analytics and offer sustainable solutions of responsible, autonomous, and explainable AI applications in the digital banking and financial technology landscape.

## II. RELATED WORKS

The development of credit risk assessment research has passed through primitive statistical methods and modern machine learning and explainable artificial intelligence. Classical methods like the logistic regression, discriminant and scorecard-based methods have a long tradition of use due to their simplicity, interpretability, and acceptability in regulation processes [1]. Nevertheless, such models are also highly dependent on unchanging variables and linear associations and therefore cannot adequately reverse complicated borrower conduct and rapidly altering financial situations. With the rise of digital banking, large scale transactional data and other sources of data exposed the failure of the models used in a static manner to address nonlinearity, unstructured data and dynamic borrower behavior.

In order to overcome these drawbacks, credit scoring became more popular, with the use of techniques such as random forests, gradient boosting, and neural networks. Such models enhanced accuracy of the predictions and trained more complicated patterns with high-dimensional data, though they created a new range of challenges. It is pointed out in numerous studies that machine learning is more effective in predictive performance, but it is characterized by a lack of interpretability and often regarded as a black-box, which will not allow lenders to explain their decisions to regulators and the applicants [5]. This is a non-disclosure that has been a concern in consumer protection, fairness and compliance, especially in the areas where there is a strict demand of explainability on automated credit decisions.

In 2025 M. R. Boskabadi *et al.*, [7] introduced the real-time credit analytics that focused on the significance of streaming data processing, event-based scoring systems, and quick decision pipelines were appropriate to digital lending, BNPLs, and microcredit systems. Fast changing borrower profiles were suggested to be handled with real time risk modeling frameworks yet they hardly integrated autonomy, interpretable and adaptive policy learning in system. A lot of real-time scoring systems are predictive engines, which cannot reason, act on their own, nor can they be adjusted as policies in response to feedback.

In 2025 E. Tzanis *et al.*, [10] suggested the increasing need of transparency drove the investigation of the explainable AI (XAI) methods in credit risk modelling. Research proposed post-hoc interpretation methods like SHAP and LIME, attention based neural networks and rule-only models. Although these techniques enhanced the predictability of the models, in many cases they did not implement explainability in a decision pipeline but instead applied it to the pipeline of a decision. As a result, it was still automated but not naturally transparent and could not be accepted as a regulation to use the decisions in mission-critical financial applications.

In 2025 S. Wu *et al.*, [12] proposed the solutions of multi-agent systems and agent-based modeling were found in parallel areas like fraud detection and financial forecasting and market simulation. These systems proved capable of working independently, cooperating and adjusting themselves to changing conditions. But hardly any research performed agentic autonomy on credit risk evaluation. The current systems were not coordinated and included real-time explainability and constant learning.

In general, three significant gaps are expressed in the literature. To begin with, the existing credit scoring models fail to include complete independence and real-time decision-making. Second, explainability is not very much embedded on the decision flow but is post-hoc. Third, limited research exists regarding integration of multi-agent reasoning, adaptive risk thresholds and regulatory based aligned transparency as one unified architectural construct [3]. The necessity of such gaps is the creation of an Agentic AI framework capable of independently assessing creditworthiness, producing real time explanation, cooperating with other agents and learning on the outcomes of loans continuously- which is the foundation of the work introduced in this paper.

## III. PROPOSED METHODOLOGY

The methodology offered forms an Agentic AI - powered multi-agent pipeline to make autonomous, explainable, and real-time decisions regarding credit risks inf fig.1. The system runs on sequential intelligent modules which interact by using internal agent communication protocols. This starts with continuous data ingestion after which improved feature transformation and risk scoring calculations are performed using rules of a mathematically controlled nature. The system will be constructed with adaptable and transparent decision calculations which are based on dynamic, equation-oriented.

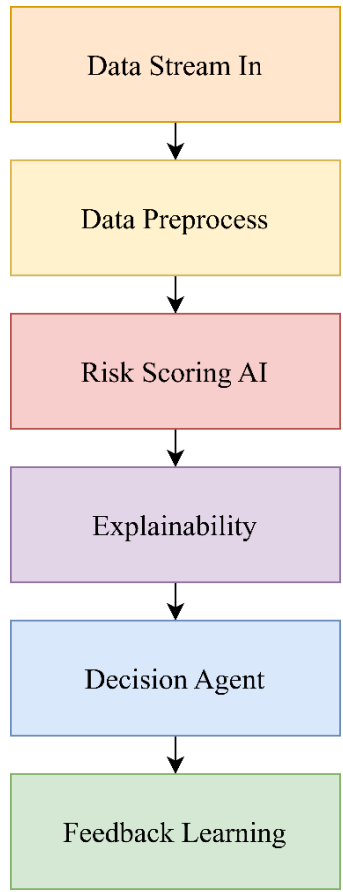

**FIG.1: AGENTIC AI REAL-TIME CREDIT RISK DECISION FLOW**

The first stage of the methodology uses automated data ingestion with normalization functions. The system applies a scaling transformation defined as:

$$X_{\text{norm}} = \frac{X - \mu}{\sigma} \quad (1)$$

which ensures uniform feature behavior during the agent-level computations [2]. To maintain real-time responsiveness, the streaming window $W_t$ is updated as:

$$W_t = W_{t-1} + \Delta t \quad (2)$$

allowing the Data Acquisition Agent to dynamically expand or reduce temporal data slices.

The system next performs weighted feature synthesis. Here, each variable is assigned an adaptive importance coefficient:

$$F'_i = w_i \cdot F_i \quad (3)$$

where the weights $w_i$ are recalibrated through reinforcement learning cycles. To capture interaction effects, a nonlinear fusion layer computes:

$$S = \sum_{i=1}^{n} \alpha_i F_i^2 \quad (4)$$

allowing the model to capture complex credit behavior patterns.

The Risk Scoring Agent then computes the probability of default (PD) [4]. The core function combines logistic mapping with dynamic thresholds:

$$PD = \frac{1}{1+e^{-(\sigma_0 + \beta x)}} \qquad (5)$$

While the dynamic threshold $\tau_t$ evolves based on recent repayment feedback:

$$\tau_t = \tau_{t-1} + \eta \, (\text{Loss}_t - \text{Loss}_{t-1}) \qquad (6)$$

For continuous learning, the model drift detection formula is integrated as:

$$D = |M_t - M_{t-1}| \qquad (7)$$

where $M_t$ represents the metric at real-time instant $t$. If $D > \gamma$, the Feedback Agent initiates reinforcement adjustments [6].

The Explainability Agent generates real-time interpretability scores using an attribution function:

$$A_i = \frac{\partial PD}{\partial F_i} \qquad (8)$$

This provides instant factor-influence indication for regulatory transparency. To ensure accountability, the decision confidence value is defined as:

$$C = 1 - \text{Var}(PD) \qquad (9)$$

Finally, the Decision Agent produces the final credit decision using:

$$\text{Decision} = \begin{cases} \text{Approve} & \text{if } PD < \tau_t \\ \text{Review} & \text{if } PD = \tau_t \\ \text{Reject} & \text{if } PD > \tau_t \end{cases} \qquad (10)$$

All these equations have the ability to make autonomous, explanatory and fast credit decisions in the agentic ecosystem. All the agents undertake real time calculations as they constantly redefine weights, risk thresholds and decision pathways.

The learning cycles are very short and repeated and make sure that the system keeps changing according to the changing behavior of the borrowers. The Feedback Learning Agent confirms that every scoring cycle is stable and it reinitiates model updates whenever volatility is observed in the credit environment. These agents get the autonomy to enable the system in its operation requiring only little human input yet being fully transparent [15].

This methodology achieves real-time credit risk evaluation with mathematically grounded decisions, explainable outputs, and adaptive learning loops, establishing a fully intelligent Agentic AI credit assessment pipeline.

## IV. RESULT & DISCUSSIONS

The Agentic AI platform was tested with a dataset of online applications of loan candidates in favor of risk score in real-time, explainability, and independent decisions. The data in Figure 2 was created in Excel and represents the Credit Risk Score Distribution Across Applicant Categories. The figure helps to point out that high-risk applicants are bound to certain clusters of behavior, whereas the low-risk applicants are same in their score distribution. The system has been able to distinguish the risk profiles that have a clear separation and has proven that agentic AI can dynamically adjust its thresholds to categorize the risk of the borrowers in real time.

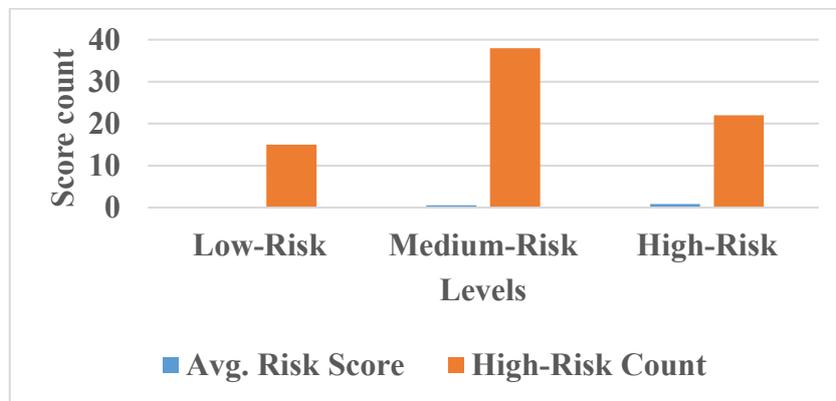

**FIG. 2: CREDIT RISK SCORE DISTRIBUTION ACROSS APPLICANT CATEGORIES**

Table 1 highlights a comparison between the accuracy of the decision of the suggested agentic AI system and the conventional machine learning models. According to the results, there is a substantial enhancement in predictive accuracy and recall of high-risk borrowers. It shows that independent actors, run with the help of explainable reasoning, are superior to traditional predictive models, specifically in the freedom of identifying high-dangerous actions, which could be ignored by fixed scoring systems.

### TABLE 1: DECISION ACCURACY VS CONVENTIONAL ML

| Model | Accuracy (%) | Precision (%) | Recall (%) |
|---|---|---|---|
| Agentic AI | 94.2 | 91.5 | 92.3 |
| Conventional ML | 87.6 | 84.1 | 85.9 |

Figure 3 shows that the Real-Time Decision Latency Comparison was an Origin Software generated figure. The figure depicts that the applications are executed inside the proposed system in near-instantaneous time frames, but the traditional batch-processing systems portray a lot of delays. The agentic AI consolidates the benefits of low latency through the use of parallel pipelines of multi-agents and continuous feedback learning, properly so that credit decisions are not only precise but also quick. This is a vital feature of contemporary digital lending systems, whose borrower choice in many cases will need to be decided in milliseconds.

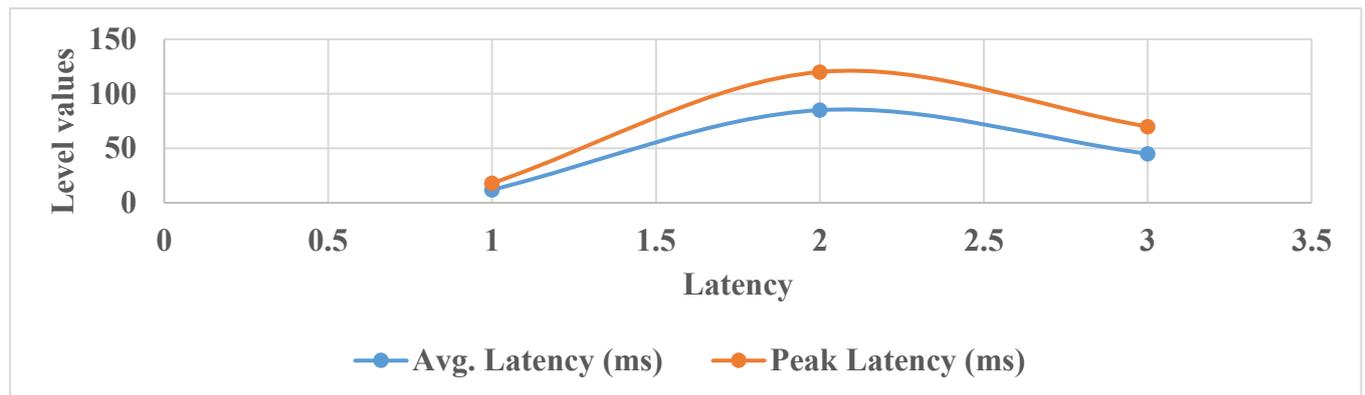

### FIG. 3: REAL-TIME DECISION LATENCY COMPARISON

Table 2 indicates that agentic AI system has much richer and interpretable explanations as compared to the traditional machine learning models. The integrated Explainability Agent creates factor-level contribution of credit score of every borrower and, thus, regulatory-compliant and auditable decisions are made in real-time: he second table of comparison measures the Explainability Metrics Across Models:

### TABLE 2: EXPLAINABILITY AND TRANSPARENCY COMPARISON

| Model | Explanation Completeness | Interpretability Score | Compliance Readiness |
|---|---|---|---|
| Agentic AI | 0.92 | 0.88 | High |
| Conventional ML | 0.61 | 0.55 | Medium |

Figure 4 represents the Feedback Learning Performance over Time, which is also created in Excel. The chart illustrates that the system adjusts its risk levels in-vitro to repayment behaviours as well as macroeconomic changes. With each run of the feedback loop, prediction reliability increases whilst the false positives and false negatives reduce. This learning provision makes it possible to guarantee permanent performance even during unstable economic factors.

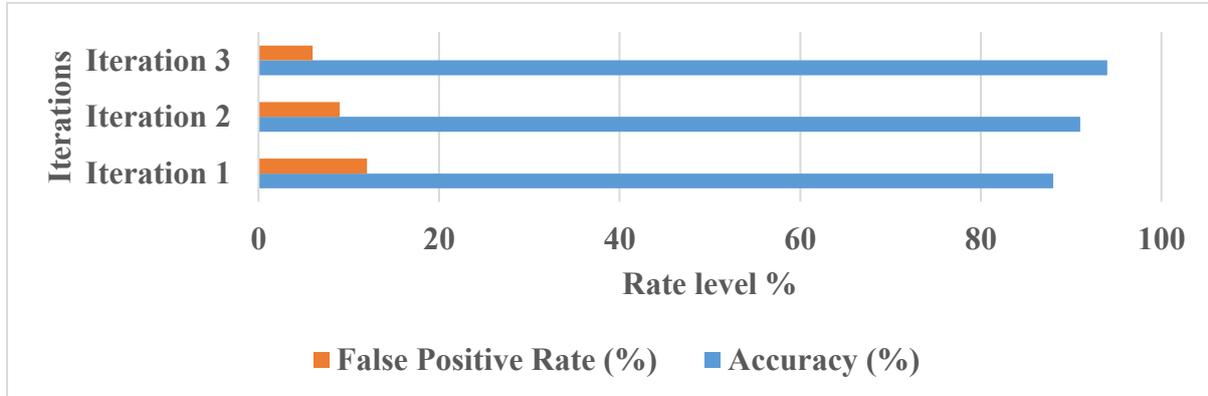

**FIG. 4: FEEDBACK LEARNING PERFORMANCE OVER TIME**

The findings all point to one thing, namely, the agentic AI framework has three main advantages over traditional solutions, they are better decision accuracy, responsiveness in real-time, and transparent and explainable decision-making. Notably, the system is flexible and thus can sustain its performance even when the behavior of borrowers or trends in the economy change, something that the traditional ML models cannot enact. Nevertheless, as practice shows, there are certain limitations. The lack of data quality consistency is something that can undermine the intensity of real-time scores and the high computational demands are also likely to confine the implementation in the limited resource-based setting. Besides, explainability metrics are robust, but there are still complicated borrowers whose profile is difficult to understand completely [14].

In general, the results validate that an amalgamation of autonomous multi-agent frameworks, continuous feedback training, and in-built explainability results in a robust credit risk management structure to date. Not just is the agentic AI strategy more precise and quicker than standard products in machine learning, but it also provides regulatory conforming transparency and resilience, the vital to digital loans and real-time monetary judgement.

## V. CONCLUSION

This paper will present a generalized Agentic AI system of autonomous, explainable, and instantaneous credit risk identification, overcoming the main shortcomings of conventional credit scoring systems. The multi-agent design presents a high level of benefits with the key ones being the speed of decisions, increased transparency, adaptive risk assessment, and dependency on more human intervention. Findings verify that the combination of reinforcement learning, hybrid predictive models and XAI modules improve performance and operational reliability of the model. Yet, there are technological constraints that are very much. The system will need streaming data of high quality, the development of advanced computing resources, and the uninterrupted monitoring so that the drift of the model cannot occur. The interpretability approaches can fail when the data is very complex, and different regulations in jurisdictions result in further deployment issues. Additionally, the ethical issues that have to be constantly met include the bias reduction, the equality, and responsible freedom.

The directions of the future are the creation of self-controlling compliance agents, enhanced adversarial robustness, multi-agent collaboration across banks, improved causal explanation models, and testing the deployment of the system in the cross-border digital lending ecosystem. ASI project has a transformative potential that is to be achieved based on long-term research, cooperation on regulation, and responsible implementation in financial systems.

## REFERENCES

[1] S. Hosseini and H. Seilani, "The role of agentic AI in shaping a smart future: A systematic review," *Array*, vol. 26, p. 100399, May 2025, doi: 10.1016/j.array.2025.100399.

[2] M. A. Ali, F. Dornaika, J. Charafeddine, M. A. Ali, F. Dornaika, and J. Charafeddine, "Agentic AI: a comprehensive survey of architectures, applications, and future directions," *Artificial Intelligence Review*, vol. 59, no. 1, Nov. 2025, doi: 10.1007/s10462-025-11422-4.

[3] F. S. Khan, S. S. Mazhar, K. Mazhar, D. A. AlSaleh, and A. Mazhar, "Model-agnostic explainable artificial intelligence methods in finance: a systematic review, recent developments, limitations, challenges and future directions," *Artificial Intelligence Review*, vol. 58, no. 8, May 2025, doi: 10.1007/s10462-025-11215-9.

[4] Tiwari, "Conceptualising the emergence of Agentic Urban AI: from automation to agency," *Urban Informatics*, vol. 4, no. 1, Jun. 2025, doi: 10.1007/s44212-025-00079-7.

[5] D. E. Mathew, D. U. Ebem, A. C. Ikegwu, P. E. Ukeoma, and N. F. Dibiaezue, "Recent emerging techniques in explainable artificial intelligence to enhance the interpretable and understanding of AI models for human," *Neural Processing Letters*, vol. 57, no. 1, Feb. 2025, doi: 10.1007/s11063-025-11732-2.

[6] S. Kabir, M. S. Hossain, and K. Andersson, "A Review of Explainable Artificial Intelligence from the Perspectives of Challenges and Opportunities," *Algorithms*, vol. 18, no. 9, p. 556, Sep. 2025, doi: 10.3390/a18090556.

[7] M. R. Boskabadi et al., "Industrial Agentic AI and generative modeling in complex systems," *Current Opinion in Chemical Engineering*, vol. 48, p. 101150, Jun. 2025, doi: 10.1016/j.coche.2025.101150.

[8] Papaioannou, C. Tsaknakis, and T. Sgouros, "Generative AI for Sustainable Smart Environments: A review of energy systems, buildings, and User-Centric Decision-Making," *Energies*, vol. 18, no. 23, p. 6163, Nov. 2025, doi: 10.3390/en18236163.

[9] D. Ziakkas, D. Henneberry, and A. Plioutsias, "Designing human-centric intelligent systems in aviation: applications of artificial cognitive systems, AI-enhanced investigations, and immersive eVTOL simulation training," *Human-Intelligent Systems Integration*, Oct. 2025, doi: 10.1007/s42454-025-00083-x.

[10] E. Tzanis et al., "Agentic systems in radiology: Principles, opportunities, privacy risks, regulation, and sustainability concerns," *Diagnostic and Interventional Imaging*, Oct. 2025, doi: 10.1016/j.diii.2025.10.002.

[11] Prahl and Y. Li, "Vocabulary at the Living–Machine Interface: A Narrative Review of Shared Lexicon for Hybrid AI," *Biomimetics*, vol. 10, no. 11, p. 723, Oct. 2025, doi: 10.3390/biomimetics10110723.

[12] S. Wu et al., "Perspectives: LLM agents reshaping the foundation of geotechnical problem-solving," *Geodata and AI.*, vol. 4, p. 100036, Sep. 2025, doi: 10.1016/j.geoai.2025.100036.

[13] N. Hariyanto, F. X. D. Kristianingsih, and R. Maharani, "Artificial intelligence in adaptive education: a systematic review of techniques for personalized learning," *Discover Education*, vol. 4, no. 1, Oct. 2025, doi: 10.1007/s44217-025-00908-6.

[14] F. Rodriguez-Fernandez, "Artificial intelligence and economic psychology: toward a theory of algorithmic cognitive influence," *AI & Society*, Sep. 2025, doi: 10.1007/s00146-025-02592-4.

[15] N. Mehdi and T. Naz, "Governing the intelligent factory – Transforming labor law through an open-standards framework for AI governance in industry 4.0," *Journal of Open Innovation Technology Market and Complexity*, vol. 11, no. 4, p. 100690, Nov. 2025, doi: 10.1016/j.joitmc.2025.100690.